\documentclass[preprint]{article}

\usepackage{microtype}
\usepackage{graphicx}
\usepackage{subcaption}
\usepackage{booktabs}
\usepackage{listings}
\usepackage{xcolor}
\usepackage{multirow}
\usepackage{siunitx}
\usepackage{enumitem}

\usepackage{hyperref}

\definecolor{codegreen}{rgb}{0,0.6,0}
\definecolor{codegray}{rgb}{0.5,0.5,0.5}
\definecolor{codepurple}{rgb}{0.58,0,0.82}
\definecolor{backcolour}{rgb}{0.95,0.95,0.92}
\definecolor{codecyan}{rgb}{0.0, 0.6, 0.6}  % Color for built-ins

\lstdefinestyle{mystyle}{
    language=Python,
    backgroundcolor=\color{backcolour},
    commentstyle=\color{codegreen},
    keywordstyle=\color{magenta}, % Standard keywords (def, for, in)
    numberstyle=\tiny\color{codegray},
    stringstyle=\color{codepurple},
    basicstyle=\ttfamily\footnotesize,
    breaklines=true,
    morekeywords={self},
    emph={print, len, range, open, str, int, float, input, type, list, dict, set, tuple},
    emphstyle=\color{codecyan}
    emph={[2]jax}
    emphstyle={[2]\color{codepurple}\bfseries}, % Blue and Bold
}

\usepackage{icml2026}

\usepackage{amsmath}
\usepackage{amssymb}
\usepackage{mathtools}
\usepackage{amsthm}

\usepackage[capitalize,noabbrev]{cleveref}

\theoremstyle{plain}

\theoremstyle{definition}

\theoremstyle{remark}

\usepackage[textsize=tiny]{todonotes}

\usepackage{xspace}
\newcommand{\jax}{JAX\xspace}
\newcommand{\torch}{PyTorch\xspace}
\newcommand{\opacus}{Opacus\xspace}
\newcommand{\jp}{\jax-Privacy\xspace}
\newcommand{\code}[1]{\texttt{#1}}

\icmltitlerunning{\jp: A library for differentially private machine learning}

\begin{document}

\twocolumn[
  \icmltitle{\jp: A library for differentially private machine learning}

  \begin{icmlauthorlist}
    \icmlauthor{Ryan McKenna}{} %{googleresearch}
    \icmlauthor{Galen Andrew}{}
    \icmlauthor{Borja Balle}{}    
    \icmlauthor{Vadym Doroshenko}{}
    \icmlauthor{Arun Ganesh}{}
    \icmlauthor{Weiwei Kong}{}
    \icmlauthor{Alex Kurakin}{}
    \icmlauthor{Brendan McMahan}{}
    \icmlauthor{Mikhail Pravilov}{}

  \end{icmlauthorlist}

  %\icmlaffiliation{googleresearch}{Google Research}

  \vskip 0.3in
]

% \begin{abstract}
% \jp is a library designed to simplify the deployment of robust and performant mechanisms for differentially private machine learning. Guided by design principles of usability, flexibility, and efficiency, \jp serves both researchers requiring deep customization and practitioners who want a more out-of-the-box experience. The library provides verified, modular primitives for critical components for all aspects of the mechanism design including batch selection, gradient clipping, noise addition, accounting, and auditing, and brings together a large body of recent research on differentially private ML.
% \end{abstract}

\section{Introduction}

The deployment of machine learning (ML) on sensitive data requires rigorous privacy guarantees to mitigate the risks of unintended memorization and membership inference. While Differential Privacy (DP) provides the necessary theoretical framework, there is a rich space of mechanisms to consider and implementation challenges to overcome to train private models effectively in practice.

Implementing private ML mechanisms in a correct and performant manner can be tricky and requires significant engineering effort. For example, one common challenge is efficiently handling per-example gradient clipping. Implementing this operation as a sequential loop can lead to poor accelerator utilization, while implementing it as a vectorized operation can lead to high memory requirements. 

Perhaps the most dangerous challenge is the issue of silent failure. Unlike standard machine learning bugs that manifest as exploding gradients or poor convergence, a flawed privacy mechanism often works \emph{too well}, as the model trains smoothly and accuracy improves, effectively masking the fact that the privacy guarantee has been violated \cite{DBLP:journals/corr/abs-2202-12219}. As a general rule, verifying an ad-hoc implementation requires careful review from DP experts.

\textbf{\jp} addresses these challenges by providing a centralized repository of robust, efficient, and verified components. By relying on a shared library rather than reimplementing mechanisms end-to-end in isolation, researchers benefit from vetted primitives for key components that are optimized for the \jax ecosystem \cite{jax2018github}. This consolidation allows developers to focus on algorithm design rather than the low-level implementation details, ensuring that the critical shared plumbing of private training is both correct and performant.

The library is designed to serve two types of users. First, it empowers researchers who require deep customization of the training loop to prototype novel mechanisms, working directly at the \jax level. Second, it supports practitioners who need ``batteries-included'' implementations to apply established privacy mechanisms to standard Keras or Flax models with minimal boilerplate code.

Guided by the design principles of usability, flexibility, and efficiency, \jp democratizes access to optimized tools for private ML. The library aims to improve research velocity and broaden access to state-of-the-art privacy mechanisms, while encouraging users to contribute back to the library. Specifically, \jp delivers:

\begin{itemize}[leftmargin=*]
\item \textbf{Usability}: We provide simple, framework-agnostic APIs that follow standard \jax and Optax patterns, and can be used across the entire \jax ecosystem. Users familiar with \jax should feel right at home using \jp.

\item \textbf{Flexibility \& Utility}: We provide the tools to implement a variety of mechanisms, providing access to state-of-the-art research going far beyond standard differentially-private stochastic gradient descent (DP-SGD).

\item \textbf{Efficiency \& Scalability}: Each component of our library has been carefully tuned and optimized.  As a result of this effort, private training loops written with \jp often have comparable throughput to their non-private counterparts. Moreover, these components work well in both single-machine and large-scale distributed scenarios.

\item \textbf{Robustness \& Correctness}: The formal guarantees of the library components are precisely documented and thoroughly tested. Edge cases like empty batches and NaN gradients are safely handled to preserve DP.
\end{itemize}

\section{Core Library}

Most mechanisms for differentially private ML consist of the following key components: \emph{batch selection}, (per-example) \emph{gradient clipping}, \emph{noise addition}, \emph{accounting}, and (optionally) \emph{auditing}. \jp provides APIs and implementations for each of these components. In this section, we describe each component including notes about the implementation and how it relates to our design philosophy. 

\paragraph{Batch Selection}

There is a major difference between the theoretical analysis of DP-SGD, which assumes i.i.d.\ Poisson sampling, and practical implementations that typically shuffle and load fixed-size batches sequentially. This substitution is commonly accepted in research scenarios as the sampling difference does not meaningfully affect utility, but invalidates the privacy guarantee \cite{chua24howprivate}. This gap has historically existed due to infrastructure constraints: random-access to large-scale datasets that do not fit in memory has not been available. \jp brings this component into scope, offering easy-to-swap implementations for correctness while ultimately deferring the deployment choice to the user.

Batch selection strategies are a rich research space, with a variety of approaches proposed in the literature that apply in different settings or have different trade-offs \cite{abadi2016deep,chua2024balls,chua2024scalable,balle2018privacy,choquette2023amplified,choquette2024near}. To ensure maximum flexibility, the interface is pure NumPy and returns only batch indices, remaining completely agnostic to the underlying dataset format. Forming batches from indices is trivial for in-memory arrays, and modern data loading libraries like \code{pygrain} can do it efficiently even for datasets residing completely on disk \cite{grain2023github}.  In addition to the challenges related to random-access, batch selection strategies typically produce variable-sized batches. Handling this naively would require recompiling the train step function for every unique batch size encountered and can be prohibitively expensive.  We therefore provide utilities for padding batches with dummy examples that do not contribute to training. These utilities can pad batches to a fixed size (or set of sizes), reducing the number of recompilations needed during training.

\paragraph{(Per-Example) Gradient Clipping}

Historically, implementing per-example gradient clipping involved a stark trade-off: sequential loops offered poor resource utilization, while naive vectorization caused memory explosions. \jp resolves this by implementing clipping via higher-order functions, specifically composing \code{jax.vmap}, \code{jax.grad}, and \code{jax.lax.scan} to introduce a \code{microbatch\_size} parameter. This design allows users to seamlessly interpolate between sequential and fully vectorized execution, effectively managing memory footprints while hiding XLA dispatch overheads. Consequently, the computational cost is often comparable to standard \code{jax.grad}, and in some regimes, microbatching enables larger batch sizes than native gradient computation.

These capabilities are exposed as drop-in replacements for \code{jax.grad} and \code{jax.value\_and\_grad}, maintaining the flexibility of \jax, while seamlessly supporting DP-specific configuration like clipping in diverse geometries or at different levels (example-level or group-level). Crucially, the returned function attaches a sensitivity property, helping to prevent bugs where noise calibration parameters diverge from the mechanism's configuration. This approach avoids the pitfalls of ``ghost clipping,'' \cite{lee2021scaling,bu2023differentially} an alternative technique that, while effective, suffers from significant drawbacks: it requires deep integration with specific neural network layers, typically necessitates two backward passes, and demands substantial expertise to verify correctness. By avoiding these constraints, \jp ensures broad compatibility and verifiable correctness.

\paragraph{Noise Addition}

Noise addition is a key component of all DP mechanisms. The simplest approach, DP-SGD, injects i.i.d.\ Gaussian noise at each iteration, a process that is effectively stateless. In contrast, more sophisticated strategies such as DP-FTRL or Matrix Factorization (DP-MF) employ correlated noise, requiring memory of previously injected noise. To accommodate this diversity, \jp implements its noise addition routines via the standard \texttt{optax.GradientTransformation} API. This design choice presents users with a simple and familiar interface, enabling them to switch between different noise generation strategies without altering the core training loop.

Within the library's \code{matrix\_factorization} module, we provide a number of methods to optimize various classes of correlated noise strategies, allowing users to configure a wide array of mechanisms established in the literature. See \citet{pillutla2025correlated} for a recent and comprehensive survey of correlated noise mechanisms; Jax Privacy was in fact used for figures and experiments in this paper.  Beyond this flexibility, the library addresses the computational demands of high-dimensional noise generation. Recognizing that noise generation is an embarrassingly parallel problem, the \jp implementation exploits this property in multi-machine scenarios. Users can scale their noise generation across distributed systems with a simple configuration option, ensuring that the noise addition step is not a bottleneck even with strategies that have high memory requirements.

\paragraph{Accounting}

Accounting refers to the computational process of quantifying privacy properties (e.g., $\epsilon$ and $\delta$) for a mechanism given its parameters or by calibrating the noise variance to achieve a desired privacy target. The importance of tight accounting cannot be overstated; improving the tightness of the privacy analysis effectively yields ``free'' utility by allowing for less noise to be added for the same formal guarantee, without any algorithmic changes. Rather than implementing a custom solution, \jp builds directly on top of the \texttt{dp-accounting} library, which provides the primitive building blocks necessary to do accounting for the mechanisms that can be written with \jp. This integration provides users with a standardized, verified foundation for tracking privacy budgets across diverse training configurations. As examples, \jp supports accounting for DP-SGD with truncated batches \cite{ganesh2025tighter}, group privacy guarantees \cite{charles2024fine}, and variants such as DP-BandMF \cite{choquette2023amplified}.

\paragraph{Auditing}

Empirical privacy auditing estimates the privacy leakage of a mechanism by simulating an adversary's attempt to distinguish between neighboring datasets (i.e., mounting a membership inference attack). It can function as a rigorous end-to-end check for differential privacy implementations: privacy auditing establishes a formal lower bound on privacy leakage, which can indicate a regression or bug in the privacy accounting or implementation if the empirically observed leakage $\epsilon_{emp}$ exceeds the theoretical guarantee $\epsilon_{theory}$~\citep{jagielski2020auditing, nasr2023tight}. Auditing also provides a practical signal regarding the effective privacy leakage. While $\epsilon_{theory}$ establishes a provable worst-case upper bound, it frequently grants the adversary unreasonable powers (e.g., the ability to observe all training checkpoints), or relies on loose numerical bounds. Empirical auditing quantifies the leakage a specific adversary (often with advantages beyond what we expect of any real-world adversary) can achieve, often revealing a substantial gap where $\epsilon_{emp} \ll \epsilon_{theory}$ \cite{jayaraman2019evaluating}. While a low $\epsilon_{emp}$ cannot prove the system is secure, as a stronger, undiscovered attack might exist, it can offer strong evidence for privacy by demonstrating that a carefully designed attack does not succeed. It is also useful for distinguishing training methods that vary in their empirical leakage, despite similar analytical guarantees \citep{hu2025empirical}.

\jp supports a suite of state-of-the-art auditing methods and metrics proposed in the literature~\cite{nasr2021adversary,steinke2023privacy,mahloujifar2024auditing}. The \texttt{auditing} component of \jp is designed for a typical auditing scenario where adversarial ``canary'' examples are crafted and inserted into the data with some probability, and scores are produced instantiating an attack that attempts to determine whether a given canary was included or excluded from training. While the design of canaries and the scoring function is highly application specific and left to the discretion of the user, \jp contains functions for estimating and bounding various metrics of the membership inference attack.

\paragraph{Keras API}

To bridge the gap between research flexibility and production readiness, JAX-Privacy includes a dedicated \texttt{keras\_api} designed for the managed, end-to-end training of JAX models defined within the Keras framework. While the Core Library offers granular control for algorithmic research, the Keras API targets practitioners who require a robust, out-of-the-box solution with minimal boilerplate. This high-level interface abstracts away the complexities of the training loop, allowing users to define a privacy-preserving model simply by providing a standard Keras model definition and data in a supported format.

In the Keras API, users specify core mechanism parameters like the privacy budget, batch size, and number of iterations through a simple configuration object. This config has many customizable parameters, but it provides reasonable default settings for most of them, reducing the cognitive load on the user. This approach prioritizes ease of use and standardization, offering a managed path to differential privacy, and although it is less flexible than the lower-level primitives, it ensures a correct and efficient implementation for standard use cases vetted by the JAX-privacy authors.
\section{Highlighted Applications}

While \jp works well in single-machine environments common in smaller-scale academic research, the core components were carefully written to work well in large-scale settings frequently encountered at Google where both the data and the model may be partitioned across many machines. Internally, we have used \jp to fine-tune multi-billion parameter models on thousands of machines with DP, and to pre-train the 1B parameter VaultGemma model with DP \cite{sinha2025vaultgemma}. \jp users can fork  our examples to fine-tune Gemma models with DP.

When using \jp in multi-machine environments, users are primarily responsible for specifying the sharding strategy for their data, model, and model forward pass, while \jp ensures the DP-specific intermediates have the expected sharding. Applying \jp at these large scales helped harden our implementations, and we documented some of the sharp edges we encountered at \texttt{https://jax-privacy.readthedocs.io}. 

While \jp is designed to work seemlessly within the JAX ecosystem, integrating it within a neural network training library is not always trivial.  Correct DP implementations require tight coupling of all components, while in existing libraries these components may be much more loosely coupled, making it difficult to both implement DP correctly and audit that an implementation is correct, while adhering to the structure and organization of the library.
\section{Related Work}

% Put this table from experiments here so it renders on the right page.
\begin{table*}[t!]
    \centering
    \caption{Throughput, measured as examples processed per second (higher is better), for training loops implemented with \jp, \jax, Opacus, and PyTorch, on standard model architectures. Relative throughput compared to \jax is shown in parentheses.}
    \label{table:experiments}
    \begin{tabular}{ll|llll}
    \toprule
    & & \textbf{\jp} & \textbf{\jax} & \textbf{\opacus} & \textbf{\torch} \\
    \midrule
    
    \multirow{3}{*}{CNN} 
    & Small  & 31837.17 (0.50x) & 63898.91 (1.00x) & 8916.34 (0.14x) & 58820.44 (0.92x) \\
    & Medium &  2466.76 (0.48x) &  5180.58 (1.00x) &  510.19 (0.10x) &  5368.72 (1.04x) \\
    & Large  &   437.03 (0.41x) &  1057.17 (1.00x) &  182.84 (0.17x) &  1032.12 (0.98x) \\

    \addlinespace
    
    \multirow{3}{*}{State Space} 
    & Small  &  1326.49 (0.99x) &  1346.26 (1.00x) &   292.56 (0.22x) &  1532.19 (1.14x) \\
    & Medium &   323.50 (1.00x) &   322.28 (1.00x) &    51.89 (0.16x) &   363.28 (1.13x) \\
    & Large  &    51.57 (0.80x) &    64.25 (1.00x) &  {\textemdash}    &    57.55 (0.90x) \\
    \addlinespace
    
    \multirow{3}{*}{Transformer} 
    & Small  &  1826.02 (0.68x) &  2690.86 (1.00x) &  1084.22 (0.40x) &  1683.29 (0.63x) \\
    & Medium &   366.50 (0.75x) &   489.33 (1.00x) &   200.84 (0.41x) &   311.75 (0.64x) \\
    & Large  &    37.21 (0.62x) &    60.16 (1.00x) &  {\textemdash}    &    37.53 (0.62x) \\
    \bottomrule
    \end{tabular}
\end{table*}

\label{sec:related}

This section reviews some prominent DP Python libraries and JAX. It provides a detailed comparison between these systems and JAX Privacy.

\textbf{TensorFlow Privacy} \cite{tfprivacy} is one of the earliest comprehensive libraries and provides tight integration with the TensorFlow and Keras ecosystems \cite{chollet2015keras}. It implements standard algorithms like DP-SGD by wrapping existing optimizers to inject gradient clipping and noise addition. It has been instrumental in deploying DP in production settings and validating the scalability of DP-SGD on large models. 
% Besides DP optimization algorithms, TFP has several utility functions to compute the privacy cost of a particular DP-SGD run. 
Recently, TFP introduced several computational improvements to their core DP-SGD implementation, including (i) fast gradient clipping algorithms \cite{kong2023unified} for a large suite of matrix multiplication layers (e.g., fully connected layers) and sparse lookup layers (e.g., embedding layers), and (ii) sparsity-preserving training \cite{ghazi2023sparsity}. 
% More specifically, (i) is implemented using functional programming paradigms while (ii) uses more object-oriented (OO) ones. TFP supports both hardware acceleration on GPUs and TPUs.
% While feature-rich, TFP's reliance on TensorFlow's graph-based execution model (historically) and the complexity of modifying the optimization loop can introduce friction for researchers requiring low-level control over the privacy mechanism.

\textbf{Opacus} \cite{opacus2021} was designed to bring DP-ML to the PyTorch community with a focus on usability and performance.
It provides a high-level API for DP training in PyTorch by attaching a \code{PrivacyEngine} to standard optimizers. Unlike earlier approaches that relied on inefficient micro-batching, Opacus utilizes \code{GradSampleModule} to hook into the backward pass of PyTorch's \code{nn.Module} layers. 
% It captures activations and back-propagated gradients, then uses vectorized linear algebra (e.g., \code{torch.einsum}) to compute per-example gradients efficiently. 
Similar to TFP, Opacus provides fast gradient clipping methods but these are only limited to matrix multiplication layers (and embedding layers that specifically use matrix multiplications instead of sparse lookup). 
% Opacus supports hardware acceleration on GPUs but has no native support for TPUs as it primarily relies on dynamic hooks that often conflict with the static graph compilation required by XLA (Accelerated Linear Algebra), the compiler for TPUs.

% While Opacus is highly effective for standard workflows, its abstraction layer can sometimes obscure the underlying primitives, making it less flexible for researching non-standard privacy mechanisms or complex accounting schemes.

\textbf{JAX} \cite{jax2018} is a framework for high-performance numerical computing that combines a NumPy-consistent API with composable function transformations. Unlike the stateful, object-oriented paradigms of PyTorch and TensorFlow, JAX adopts a purely functional programming model. This design choice fundamentally alters how differentially private mechanisms (specifically gradient clipping and noise addition) are implemented and optimized. JAX requires functions to be pure (side-effect free), meaning all state such as model parameters, optimizer state, and PRNG keys must be passed explicitly. This contrasts with the stateful ``Privacy Engine'' objects in Opacus or TensorFlow Privacy, which attach to optimizers and mutate internal state. For DP research, JAX's explicit state management makes the flow of sensitive data and random noise transparent, significantly easing the burden of auditing and mathematical verification.

% At its core, JAX leverages XLA  to compile Python functions into optimized machine code kernels. This is critical for DP-SGD, where the primary bottleneck is the materialization of per-example gradients; in standard execution (eager mode), computing gradients for a batch of size $B$ and model size $P$ requires instantiating a tensor of shape $B \times P$, which often exceeds GPU memory limits. JAX's \code{jit} compilation enables \textit{operator fusion}, allowing the compiler to generate kernels that compute, clip, and accumulate gradients in a single pass without fully materializing the intermediate $B \times P$ tensor in high-bandwidth memory (HBM).

% The most distinct feature of JAX for privacy research is \code{vmap} (vectorizing map). In other frameworks, accessing per-example gradients requires either inefficient micro-batching (processing one example at a time) or complex implementations involving backward hooks and \code{einsum} operations, as seen in Opacus. JAX simplifies this by allowing the user to write the gradient logic for a \textit{single} example and strictly lifting it to a batch using \code{vmap(grad(loss\_fn))}. This not only simplifies the codebase but allows the XLA compiler to automatically determine the optimal vectorization strategy for the specific hardware backend (GPU or TPU).

\section{Performance Experiments}

We compare the performance of \jp with plain \jax and plain PyTorch (without DP), and to Opacus. We consider three standard model architectures: (1) CNN applied to image data, (2) State Space models applied to sequence data, and (3) Transformer models applied to sequence data. For each model, we consider 3 sizes: small ($\sim$1M parameters), medium ($\sim$10M parameters), and large ($\sim$100M parameters). In all cases, we simulate 50 iterations of training on dummy data using the \code{AdamW} optimizer. We run experiments for varying batch sizes, ranging from 1 to the maximum batch size supported by the hardware, considering values that are a power of two.  We measure the throughput, defined as the ratio of total examples processed divided by total time taken, and report the maximum throughput across all batch sizes evaluated. Interestingly, this does not always correspond to the maximum batch size that runs, although larger batches do tend to exhibit higher throughput in general. Our experiments focus on the single-machine training scenario, and were conducted on an Nvidia GeForce RTX 3060 GPU with 12 GB of HBM.  For \jp and Opacus, we use the default configurations.

\paragraph{Findings}

In \cref{table:experiments} we see that the throughput between the non-private baseline and the DP counterpart varies significantly between model class, model size, and programming framework (\jax vs.\ PyTorch).  In the best cases, the throughput of DP training with \jp nearly matches the throughput of standard training (e.g., medium and large state space models).  For CNNs, the overhead can be over $2\times$, while for Transformers, it is somewhere in between.  While this overhead is non-trivial, we note it is far lower than has been historically by other libraries for DP ML like TensorFlow Privacy, where the overhead has been much larger. Indeed, the overhead can be much larger with PyTorch/Opacus, exceeding $10\times$ in some cases.

Our implementations of these models are fairly basic, and there are a number of implementation details that can and do impact these throughput numbers in practice that are not varied in our experiments. Notably, the implementation of multi-headed self-attention, the rematerialization strategy, the dtype precision, the sharding strategy in distributed training scenarios, etc.

\newpage
\bibliography{refs}
\bibliographystyle{icml2026}

\newpage
\appendix
\onecolumn
\section*{Acknowledgements}

We would like to thank the the additional open source contributors to \jp, which at the time of writing includes Debangan Ghosh, Chaitanya Mishra, Neeraj Pathak, Mahesh Thakur, and Amol Yadav.

\section{\jp vs. Opacus vs. Tensorflow Privacy}

In \cref{sec:related} we provided a highlighted some similarities and differences between different open source software libraries for differentially private machine learning.  We now give a more detailed comparison between Tensorflow Privacy, Opacus, and \jp in \cref{tab:library_comparison}.

\begin{table*}[h]
\caption{Comparison of major differentially private machine learning libraries. \jp distinguishes itself through a purely functional design that leverages XLA for seamless compilation on TPUs and GPUs, avoiding the complexity of backward hooks used in PyTorch-based solutions.}
\label{tab:library_comparison}
\vskip 0.15in
\begin{center}
\begin{small}
\begin{tabular}{lccc}
\toprule
Feature & TensorFlow Privacy & Opacus & \jp \\
\midrule
\textbf{Base Framework} & TensorFlow / Keras & PyTorch & JAX \\
\textbf{Programming Model} & Mixed (OO \& Functional) & Object-Oriented & Purely Functional \\
\textbf{Mechanism Logic} & Optimizer Wrapper & \code{nn.Module} Hooks & \code{vmap} \& \code{grad} transforms \\
\textbf{Per-Example Gradients} & Micro-batching / Fast Clip & Hooks + \code{einsum} & Automatic Vectorization \\
\textbf{Memory Efficiency} & XLA Optimization & Virtual Steps (Accumulation) & XLA Fusion (No Materialization) \\
\textbf{Hardware Support} & GPU \& TPU & GPU (No Native TPU) & GPU \& TPU \\
\textbf{Primary Abstraction} & \code{DPKerasOptimizer} & \code{PrivacyEngine} & Modular Primitives \\
\bottomrule
\end{tabular}
\end{small}
\end{center}
\vskip -0.1in
\end{table*}
\section{Example}

\cref{fig:example} gives an example training loop written in \jp. This example implements a very basic version of DP-SGD with Poisson sampling, assuming the \code{loss\_fn} and initial params are provided by the user. The code the user is responsible for writing is quite simple, with the core library doing a lot of the heavy lifting, surfacing the key functionality through easy-to-use API surfaces.

\begin{figure*}[h]
\begin{lstlisting}[language=Python]
grad_fn = jax_privacy.clipped_grad(loss_fn, l2_clip_norm=1.0)
privatizer = jax_privacy.noise_addition.gaussian_privatizer(
  noise_multiplier * grad_fn.sensitivity()
)
strategy = jax_privacy.batch_selection.CyclicPoissonSampling(sampling_prob=0.01, iterations=1000)
optimizer = optax.adamw(1.0)
params = ...


@jax.jit
def update_fn(params, batch, noise_state, opt_state):
    clipped_grad = grad_fn(params, batch)
    noisy_grad, noise_state = privatizer.update(clipped_grad, noise_state)
    updates, opt_state = optimizer.update(noisy_grad, opt_state)
    params = optax.apply_updates(params, updates)
    return params, noise_state, opt_state

opt_state = optimizer.init(params)
noise_state = optimizer.init(params)
for batch_idx in strategy.batch_iterator(USERS):
   batch = data[batch_idx]
   params, noise_state, opt_state = update_fn(params, batch, noise_state, opt_state)
\end{lstlisting}
\caption{\label{fig:example} An example training loop written in \jp.}
\end{figure*}

\end{document}